\pdfoutput=1

\documentclass[11pt]{article}

\newcommand{\framework}{\textsc{LM-Pub-Quiz}}

\usepackage[]{acl}

\usepackage{times}
\usepackage{latexsym}

\usepackage[T1]{fontenc}

\usepackage[utf8]{inputenc}

\usepackage{microtype}

\usepackage{inconsolata}

\usepackage{graphicx}

\usepackage[inline]{enumitem}
\usepackage{makecell}
\usepackage{todonotes}
\usepackage{booktabs}
\usepackage{listings}

\definecolor{javared}{rgb}{0.6,0,0} 
\definecolor{javagreen}{rgb}{0.25,0.5,0.35} 
\definecolor{javapurple}{rgb}{0.5,0,0.35} 
\definecolor{javadocblue}{rgb}{0.25,0.35,0.75} 
 
\lstdefinelanguage{json}{
  keywords={},
  morecomment=[l]{//},
  morecomment=[s]{/*}{*/},
  morestring=[b]',
  morestring=[b]",
  sensitive=true
}

\usepackage{tabularx}

\title{\framework{}: A Comprehensive Framework for Zero-Shot Evaluation of Relational Knowledge in Language Models}

\author{Max Ploner\footnotemark[1] \and Jacek Wiland\footnotemark[1] \and Sebastian Pohl\footnotemark[1] \and Alan Akbik
  \vspace{0.6em} \\
  Humboldt Universität zu Berlin \\
  Science Of Intelligence \vspace{.2em}\\
  \texttt{<first name>.<last name>@hu-berlin.de} \\
}

\begin{document}
\maketitle

\begingroup\def\thefootnote{*}\footnotetext{Equal contribution}\endgroup

\begin{abstract}

Knowledge probing evaluates the extent to which a language model (LM) has acquired relational knowledge during its pre-training phase.
It provides a cost-effective means of comparing LMs of different sizes and training setups and is useful for monitoring knowledge gained or lost during continual learning (CL). 
In prior work, we presented an improved knowledge probe called~\textsc{BEAR} \citep{wilandBEAR2024}, which enables the comparison of LMs trained with different pre-training objectives (causal and masked LMs) and addresses issues of skewed distributions in previous probes to deliver a more unbiased reading of LM knowledge.
With this paper, we present \framework{}, a Python framework and leaderboard built around the BEAR probing mechanism that enables researchers and practitioners to apply it in their work. It provides options for standalone evaluation and direct integration into the widely-used training pipeline of the Hugging Face \textsc{transformers} library. Further, it provides a fine-grained analysis of different knowledge types to assist users in better understanding the knowledge in each evaluated LM. We publicly release \framework{} as an open-source project.

\end{abstract}

\section{Introduction}

\begin{figure}[t!]
    \includegraphics[width=\linewidth]{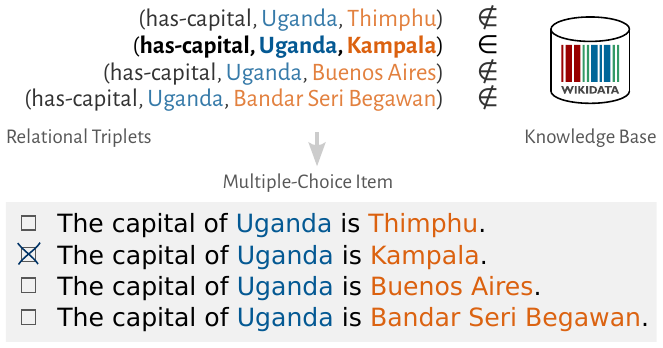}
    \caption{The BEAR probe uses relational triplets from a knowledge base (KB) to construct multiple-choice items. Here, it leverages the knowledge that ``Kampala'' is the capital of ``Uganda'', while ``Thimpu'', ``Buenos Aires'' and ``Bandar Seri Begawan'' (other \textit{capital cities}) are not. It measures whether the LM correctly ranks the verbalization of the true fact higher than the distractors.}\label{fig:bear-illustration}
    \vspace{-3mm}
\end{figure}

Pre-trained language models (LMs) currently take on a central role in state-of-the-art NLP approaches~\cite{devlinBERTPretrainingDeep2019b}. Given their importance, prior work has sought to measure the amount of factual knowledge encoded in LMs using \textit{knowledge probing} mechanisms~\citep{petroniHowContextAffects2020, kaloKAMELKnowledgeAnalysis2022}. Here, the knowledge represented in the parameters of an LM is automatically compared to factual knowledge in a relational knowledge base (KB). For instance, a probe might measure if an LM can correctly recall the capitals of countries, as illustrated in Figure~\ref{fig:bear-illustration}.

In previous work, we introduced a new knowledge probe called \textsc{BEAR}~\citep{wilandBEAR2024} that addresses various issues of ambiguities and skewed answer distributions of prior probes to deliver a more unbiased reading of LM knowledge. Further, it reformulates probing as a ranking task, thus enabling a direct comparison of LMs trained with different pre-training objectives (masked and causal LMs) and vocabularies. However, despite being conceptually simple, \textsc{BEAR} relies on a different implementation than existing probes and previously returned only an overall score as the evaluation result, thus limiting adoption and interpretability. 

\paragraph{Framework.} With this paper, we present \framework{}, an open-source Python framework and leaderboard built around the BEAR probing mechanism that enables researchers and practitioners to apply it in their work. Our framework was designed for ease of use, providing simple interfaces and direct integration into the Hugging Face \textsc{transformers} ecosystem~\cite{wolfTransformersStateoftheArtNatural2020}. Two use cases in particular have shaped the development of the library: 

\begin{enumerate}
    \item The first main use case is to evaluate and compare already-trained LMs. Users need only pass the string identifier of one of the LMs on the Hugging Face model hub in order to calculate the BEAR score for this model. This yields not only an overall BEAR score but also a more fine-grained analysis of different types of relational knowledge in the LM. 

    \item The second main use case is to monitor the knowledge gained and lost during  pre-training and continual training  (e.g.~when adapting an LM to a new domain). Here, \framework{} provides an easy integration into the Hugging Face \texttt{Trainer} to track knowledge development during training.
\end{enumerate}

To encourage uptake, we make our library freely available and open source. Additionally, we are actively curating a leaderboard with scores of existing LMs. We encourage the community to participate in extending the list of evaluated models.\footnote{The leaderboard and GitHub repository are available at \url{https://lm-pub-quiz.github.io}. Released under the MIT License.}

\section{Framework Overview}\label{sec:framework-overview}

We give an overview of \framework{}, describe how it can be installed (\ref{sec:package-setup}), explain the basic components of the interface (\ref{sec:package-interface}), and offer examples to illustrate its usage (\ref{sec:direct-eval}, \ref{sec:monitoring-knowledge}, \& \ref{sec:analysis}).

\subsection{Setup}\label{sec:package-setup}

The package containing \framework{} can be installed in the desired environment using pip:

\begin{verbatim}
  pip install lm-pub-quiz
\end{verbatim}

It relies on the \textsc{transformers} package, which users can use to load pre-trained models locally or from the Hugging Face hub.

\begin{figure}[t]
    \centering
    \begin{lstlisting}[caption={Example snippet for performing the BEAR probe on the GPT-2 model~\cite{radfordLanguageModelsAre2019}.},label={lst:direct-example},captionpos=b,basicstyle=\ttfamily\footnotesize,belowskip=-1\baselineskip]{Direct evaluation}
from lm_pub_quiz import Dataset, Evaluator

# Step 1: Load the BEAR probing dataset
dataset = Dataset.from_name("BEAR")

# Step 2: Load the LM (here: "gpt2") and create the evaluator
evaluator = Evaluator.from_model(
    "gpt2",
    model_type="CLM",
    device="cuda:0"
)

# Step 3: Run the evaluation and save the results
evaluator.evaluate_dataset(
    dataset,
    template_index=0,
    save_path="gp2_results", 
    batch_size=32,
)
    \end{lstlisting}
\end{figure}

\subsection{Interface}\label{sec:package-interface}

Our API consists of three types of objects.

\paragraph{\texttt{Dataset}} represents the dataset used to evaluate the LM.
Each dataset consists of a set of relations represented by the \texttt{Relation} class.

These relations are typically derived from the relations in the knowledge base (see Figure~\ref{fig:bear-illustration}). Relations group instances of a similar type (e.g., relation \texttt{P36} links a country or other entity to its governmental seat) and have a common set of possible answers (i.e., the options available in each multiple-choice question) and templates that are used to create the textual statements.

Each relation contains an instance table and information about their answer space.
Relations can be annotated with additional information, such as the domains of knowledge they contain and their \textit{cardinality}. By cardinality, we refer to the number of subjects for which a particular object is the correct answer: Either the relation is a one-to-one relationship or there are multiple subjects with the same answer. If the cardinality is not provided in the metadata, it is derived from the relation data.

\paragraph{\texttt{Evaluator}} is the functional component used to evaluate the model. It is instantiated with a model name (or model object). To evaluate the model on the dataset with a \texttt{Dataset}, the \texttt{evaluate\_dataset} method is called (see~\ref{sec:direct-eval}).

\paragraph{\texttt{DatasetResult}} is an object that is returned by the \texttt{evaluate\_dataset} method. This object can be used to analyze the results of a specific model. It allows the accumulation of results across the relations (e.g.~based on domains or cardinality) and enables accessing the instances-specific predictions.

\subsection{Direct Evaluation of a Trained LM}
\label{sec:direct-eval}

The first main use case of \framework{} is to evaluate the knowledge contained in a trained LM. We illustrate how to perform such an analysis for the GPT-2 model in Listing~\ref{lst:direct-example}. 

As the code example shows, it consists of three main steps:  In Step~1, we load the BEAR evaluation dataset. In Step~2, we load the LM using its string identifier on the Hugging Face model hub (here: \texttt{gpt2}) and create an evaluator for causal language models (by passing \texttt{model\_Type="CLM"}). Finally, in Step~3, we run the BEAR probe and store the evaluation results at the specified \texttt{save\_path}.

By default, all instance-level predictions are stored on the file system, allowing the computation of all metrics supported in \framework{} separately (see~\ref{sec:analysis}). It also allows for fine-grained inspection of all answers given by the LM. A more memory-efficient alternative is not to store the instance-level predictions and compute the metrics directly. This can be set by passing the \texttt{metrics} keyword to the \texttt{evaluate\_dataset} method. 

\subsection{Monitoring Knowledge during Training}
\label{sec:monitoring-knowledge}

\begin{figure}[t!]
    \includegraphics[width=\columnwidth]{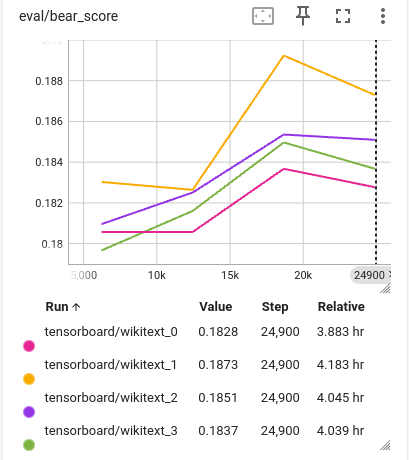}
    \caption{Example screenshot from \textsc{TensorBoard}, showcasing the Hugging Face \texttt{Trainer} integration of \framework{}. Here, we monitor the knowledge of 4 \texttt{roberta-base} models~\cite{liuRoBERTaRobustlyOptimized2019}, continuously pretrained on permutations of the Wikitext corpus.}\label{fig:tensorboard}
\end{figure}

The second use case of \framework{} is for monitoring the knowledge development in an LM during (continual) pre-training. To this end, we developed a Hugging Face \texttt{Trainer} integration. \framework{} provides a callback that can be attached to the \texttt{Trainer} instance. 
The callback will then invoke the \texttt{Evaluator} in the specified frequency.
This allows integration into monitoring tools like \textsc{TensorBoard}. See Figure~\ref{fig:tensorboard} for an illustration.

\subsection{Analysis Options}
\label{sec:analysis}

The BEAR probe consists of 60 relations retrieved from the WikiData knowledge base. Each relation connects exactly two entities to form a relation triplet. Example relations in BEAR are \textsc{has-capital} (see Figure~\ref{fig:bear-illustration}) that connects a country to its capital city, \textsc{born-in} that connects a person to their country of birth, and \textsc{crosses-river} that connects a named bridge to the river it crosses. 

Each relation in BEAR has a number of relation instances, i.e.,~specific triplets such as (\textsc{has-capital}, Uganda, Kampala). In total, BEAR has 7,731 and 40,916 of such triplets in its default and expanded variants respectively. As Figure~\ref{fig:bear-illustration} shows, each triplet is used to form one multiple-choice item question in our evaluation. The default BEAR score is the accuracy across all questions.

\framework{} offers several options for users to obtain more fine-grained analysis: 

\begin{itemize}
\item
First, users can compute separate BEAR scores for different domains of knowledge. To enable this analysis, we manually annotated each of the relations in BEAR with one or more domains (in practice, up to three), that the relations relate to.\footnote{This annotation can be found in the dataset repository: \url{https://github.com/lm-pub-quiz/BEAR}} This allows analysis of per-domain knowledge gained or lost during training.

\item
Second, one can calculate separate scores for relations based on their cardinality, as BEAR includes both 1-1 relations and 1-N relations, where the latter has multiple possible answers as opposed to just a single one. 

\item
The third option is to only aggregate the scores on a relation level. Since instances in a relation share a template, the relation-level scores reveal issues with the verbalization of the triplets.

\item Finally, one can choose to not aggregate at all, and compute the predictions per instance. This can be useful for fine-grained qualitative analysis to find knowledge bottlenecks. 
\end{itemize}

As shown in Listing~\ref{lst:retrieving-results}, the \texttt{DatasetResults} can compute these aggregated metrics. The \texttt{accumulate} keyword of the \texttt{get\_metrics} method controls the manner of aggregation and may be set to \texttt{domain}, \texttt{cardinality} or \texttt{False} for the above-mentioned aggregations.
To inspect the instance-level predictions one can use the \texttt{instance\_table} attribute of each of the \texttt{RelationResult} objects.

\begin{figure}[t]
    \centering
    \begin{lstlisting}[caption={Example of retrieving evaluation results by domain. This provides different BEAR scores to relations from domains such as "Arts", "Biographical", "Economic", etc.},label={lst:retrieving-results},captionpos=b,basicstyle=\ttfamily\footnotesize,belowskip=-1.5\baselineskip]{Retrieving the results}
from lm_pub_quiz import DatasetResults

bear_results = DatasetResults.from_path(
    "gp2_results",
    relation_info="./relation_info.json"
) 

# Get accuracy by relation types
print(bear_results.get_metrics(
    ["accuracy"], accumulate="domains"))

# Output:
#               accuracy  support
# domains                        
# Arts          0.105263   1368.0
# Biographical  0.158820   2028.5
# Economic      0.115152    770.0
# ...
    \end{lstlisting}
\end{figure}

\subsection{Comparison with Existing Libraries}

The \textsc{LM Evaluation Harness} framework \cite{eval-harness} is one of the most well-known evaluation tool for large language models, featuring numerous benchmarks as part of its task suite including knowledge tasks such as MMLU \cite{hendryckstest2021}.
This framework primarily focuses on autoregressive language models and lacks support for masked language models (MLMs).
This limitation limits the ability to compare the performance of different types of models on the same datasets.

Similarly, \textsc{HELM} \citep{liang2023holisticevaluationlanguagemodels} and \textsc{LLM-facteval} \citep{luo2023systematic} rely on the capability of CLMs to generate continuations to a prompt and are therefore not applicable to MLMs.

The unique feature of \framework{} is its focused approach to cloze statement filling, allowing the answer to appear anywhere within a sentence. This method is compatible with any type of model (whether CLM or MLM) and any tokenization. By evaluating the log-likelihood score of the entire statement instead of just its continuation or the single answer token, \framework{} overcomes the limitations of traditional \texttt{[MASK]}-predict approaches~\cite{petroniLanguageModelsKnowledge2019} without relying on text-continuation capabilities.

\section{Example Experiments}\label{sec:experiments}

To showcase example applications, we present three novel experiments that show how 
how \framework{} can be used to do conduct a detailed analysis of knowledge in the LM (Section~\ref{sec:domain-knowledge} and~\ref{sec:model-bias}) and how the Hugging Face integration (see Section~\ref{sec:cp-experiment}) can be used to monitor knowledge in a continual pre-training setting.

\subsection{Domain-specific Knowledge after Training on Different Corpora}\label{sec:domain-knowledge}

\begin{figure*}[t!]
    \centering
    \includegraphics[width=\linewidth]{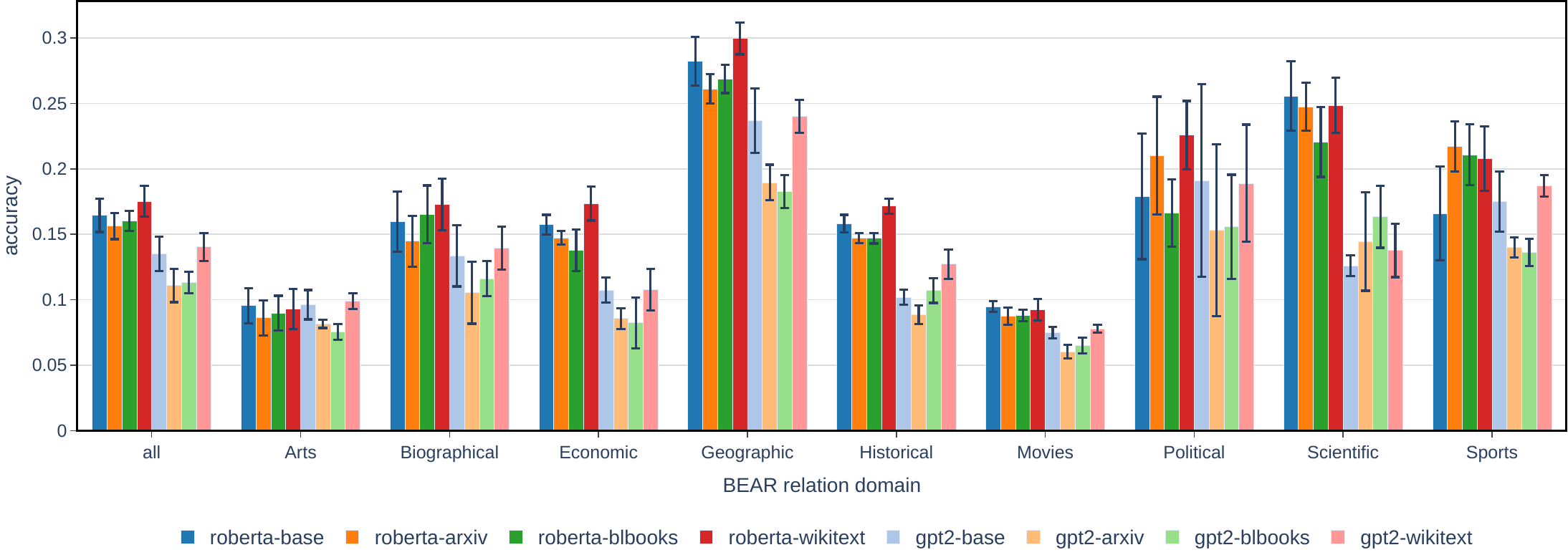}
\vspace{-6mm}
    \caption{Scores for different \texttt{BEAR} domains for models trained on different corpora.}\label{fig:bear_domain_comparison}
\end{figure*}

When adapting an LM to a specific domain, one may be interested in the various areas of knowledge contained in the model's parameters. While the overall accuracy on the complete BEAR dataset reflects the model's general knowledge, a more granular examination of the relations can provide insights into the specific areas they relate to.

\subsubsection{Experimental Setup}

We adapt two base models, \texttt{roberta-base} and \texttt{gpt2} to three domains: arXiv abstracts~\cite{clement2019arxiv}, literary texts from \texttt{blbooks} \cite{bBritishLibraryBooks2021}, and Wikipedia text from \texttt{wikitext-103-v1} \cite{merity2016pointer}. Additional information on the training setup can be found in Appendix \ref{sec:continued-domain-specific-knowledge}.

This yields a total of 8 models to compare: The two base models, the three domain-adaptations of   \texttt{roberta-base} (\texttt{roberta-arxiv}, \texttt{roberta-blbooks}, \texttt{roberta-wikitext}), and the three domain-adaptations of \texttt{gpt2-base} (i.e. \texttt{gpt2-arxiv}, \texttt{gpt2-blbooks}, \texttt{gpt2-wikitext}).

\subsubsection{Results}

Figure~\ref{fig:bear_domain_comparison} presents the analysis of all 6 models across 10 BEAR domains. We generally find that all models score highest on geographical questions and lowest on questions from the "arts" and "movies" domains. 

We also note that training with \texttt{wikitext} data improves the BEAR score the most, given that the BEAR probe was constructed from Wikidata. Further, we observe that training \textsc{GPT2} on arXiv abstracts leads to significant improvements on the scientific domain (see \texttt{gpt2-arxiv} vs \texttt{gpt2-base} in Figure~\ref{fig:bear_domain_comparison}). Further, we that the roberta-base model benefits from training on \texttt{blbooks} for the biographical and sports domains.

\subsection{Investigating Model Biases}\label{sec:model-bias}

During pre-training, models are likely to acquire various biases, primarily due to the data they were trained on~\cite{opiniongpt2024}, potentially leading them to disproportionally favor certain answers.

In this experiment, we use \framework{} with the BEAR probe to aggregate all the predicted answers given by an LM in a particular relation. Because the BEAR answer space is balanced, this aggregation results in an estimation of the model's bias, as each answer should be equally likely. We measure if different models are biased towards certain answers.

\subsubsection{Experimental Setup}
We select a single relation from the \textsc{BEAR} probe, \texttt{P30}. This relation connects locations and geographic entities to the continents they are located on (see Figure~\ref{fig:bias_p30}). 

We evaluate three pre-trained models on this relation: \texttt{roberta-base} \cite{liuRoBERTaRobustlyOptimized2019}, \texttt{gpt2} \cite{radfordLanguageModelsAre2019} and \texttt{Mistral-7B-v0.1} \cite{jiang2023mistral7b}, i.e.~one MLM and two CLMs. Model biases are estimated by applying the softmax function to the \textsc{BEAR} pseudo-log-likelihood scores, resulting in values that can be interpreted as probabilities. These values indicate the likelihood that a sentence is correct, given that at least one of the answers is correct for the given subject. Subsequently, these distributions are averaged over all subjects, resulting in the overall bias. Since each answer occurs with equal frequency for this relation, a perfect model scoring all template instances correctly would produce a uniform bias.

\subsubsection{Results}
\begin{figure}[t!]
    \includegraphics[width=\columnwidth]{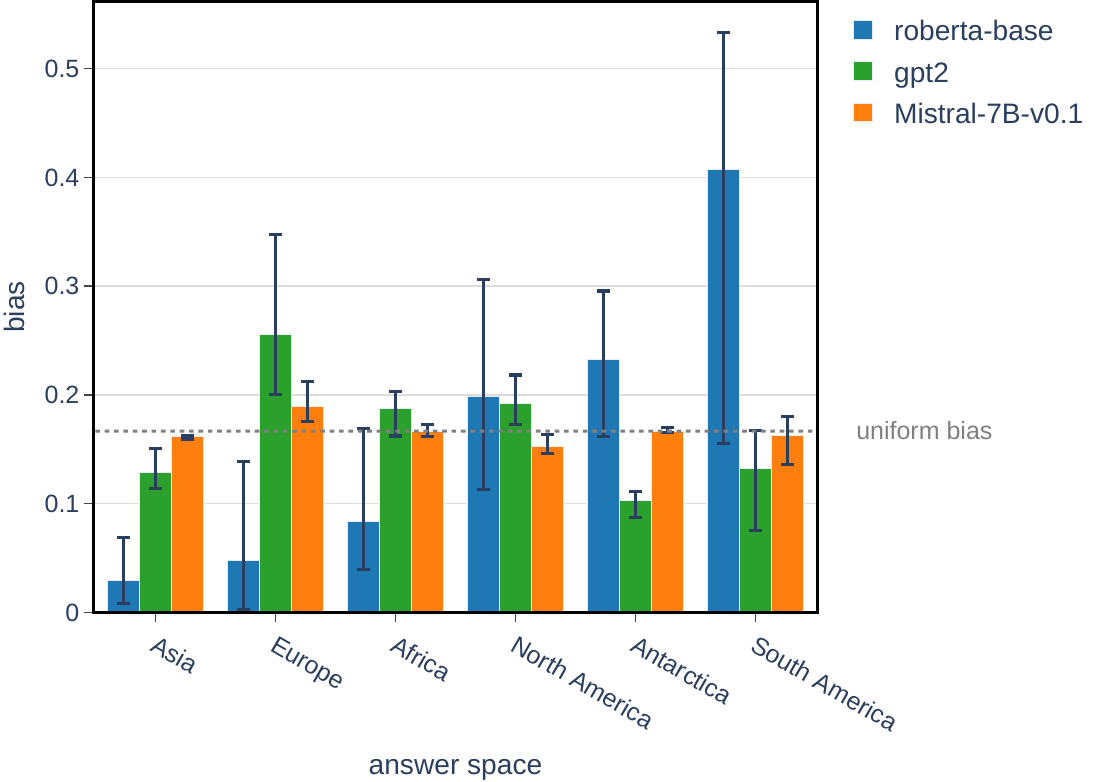}
    \caption{Performance of the selected models on the \texttt{P30} relation of the \textsc{BEAR} probe averaged over relation templates, including min and max values as bars.}\label{fig:bias_p30}
\end{figure}

Figure~\ref{fig:bias_p30} presents models' biases for relation \texttt{P30}, and shows that while \texttt{roberta-base} is biased towards the `South America' and `Antarctica' answer options, \texttt{GPT2} and \texttt{Mistral-7B-v0.1} are more likely to predict `Europe'.

Averaging over all three of its templates, relation P30 gives an accuracy of 0.98, 0.62, and 0.45 for \texttt{Mistral}, \texttt{GPT2}, and \texttt{roberta-base}, respectively. Since \texttt{Mistral-7B-v0.1} predicts most of the answers correctly, the relative answer frequency is much closer to the uniform distribution.

\subsection{Monitoring of Knowledge during Continual Learning}\label{sec:cp-experiment}

Catastrophic forgetting \citep{mccloskey1989catastrophic}, a significant challenge in continual learning, occurs when a model loses previously acquired knowledge after being trained on new datasets. We hypothesize that traditional knowledge evaluation methods such as LAMA \citep{petroniLanguageModelsKnowledge2019} employing a \texttt{[MASK]}-predict approach, overestimate the extent of forgetting of relational knowledge.

\subsubsection{Experimental Setup}

We continually train a \texttt{bert-base-cased} model using the original MLM objective on a stream of five \textit{experiences} with each experience consisting of scientific abstracts \citep{geigerArXiVArchiveTidy2019} from two scientific domains (i.e.~10 domains in total).\footnote{For an overview of the dataset and the hyper-parameters used in these experiments, see Appendix~\ref{sec:cp-experiment}.} At the end of each epoch, we evaluate the model's performance on BEAR and T-REx \citep{elsaharTRExLargeScale2018}, which is part of the LAMA benchmark.

Two methods for knowledge evaluation were used: \texttt{[MASK]}-token filling and a multiple-choice question format with a closed answer space implemented via the \framework{} package.\footnote{Due to the multi-token nature of this dataset, the [MASK]-predict method was not applicable. For a discussion of this issue, see \citet{wilandBEAR2024}.}

All scores are calculated relative to the original performance (in the pre-trained state), showing the performance change during continual pre-training. Since the \texttt{[MASK]}-filling method predicts a token over the entire vocabulary of the model (in the case of \texttt{bert-base-cased}, it is over 30K vocabulary tokens), it is inherently more difficult than choosing from a limited answer space as in \framework{}. Hence, relative scores are more suitable.

\subsubsection{Results}

\begin{figure}[t!]
    \includegraphics[width=\linewidth]{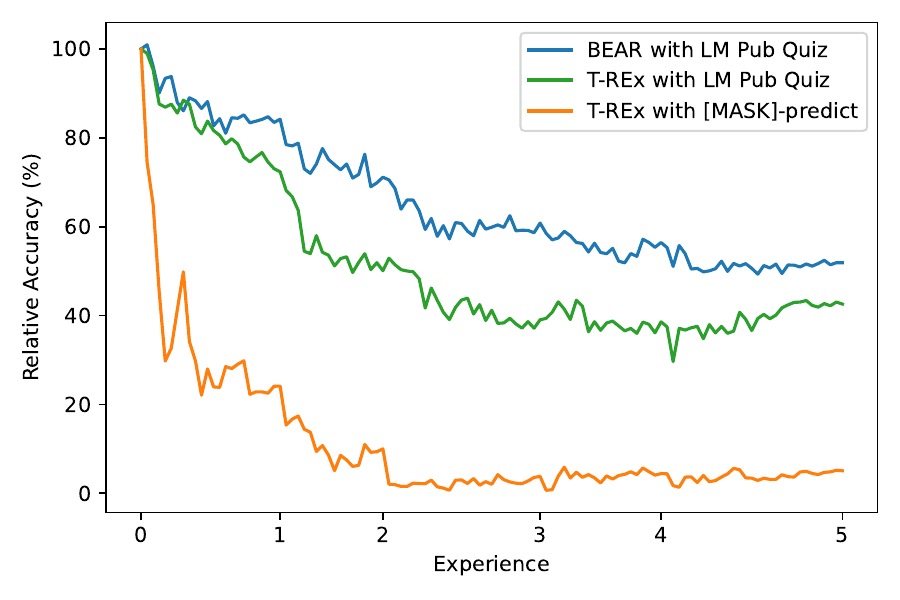}
    \caption{Trajectories of the knowledge represented in a \texttt{bert-base-cased} model throughout the continual learning process as measured by \framework{} and \texttt{[MASK]}-predict on T-REx dataset. Additionally, the performance of \texttt{bert-base-cased} evaluated on the BEAR probe is shown.}\label{fig:trex_lm_pub_quiz_mask}
\end{figure}

The forgetting curves displayed in Figure~\ref{fig:trex_lm_pub_quiz_mask} reveal the forgetting dynamics during the continual pre-training process. 
See Table~\ref{tab:continous_pretrainig_experiment2} for a detailed summary of the relative performance scored using both evaluation techniques and Section~\ref{sec:cl-experiment-additional} for additional discussion (both in the appendix).

The results indicate different performance trajectories depending on the evaluation method used. The \texttt{[MASK]}-predict approach measures a much larger degree of forgetting.  In a qualitative error analysis, we found that the model's predictions, although contextually reasonable, often do not match the expected answers due to the data distribution shift. For example, after five experiences of continual pre-training on scientific abstracts, the top three predictions for the cloze statement ``The native language of Marie Curie is \texttt{[MASK]}'' are ``considered,'' ``discussed,'' and ``presented''. While these are not necessarily incorrect in some contexts, they do not align with the expected answer ``Polish'' in T-REx.
We, therefore, believe that the \texttt{[MASK]}-filling approach may not reliably indicate the amount of relational knowledge contained within the model's parameters.

By design, \framework{} only considers answers that are appropriate to the context (but, except for one, not factually correct).
With this approach, there is a much smaller decrease in performance, especially after the first experience. This characterization of catastrophic forgetting aligns with other research on continual learning, such as \citet{cossuContinualPreTrainingMitigates2022a}, which found that unsupervised and self-supervised training objectives can partially mitigate the problem of forgetting in sequential learning.

The evaluation results on BEAR reveal a forgetting behavior similar to that observed in T-REx. However, the performance degradation observed with the BEAR probe is notably the least severe among all the experiments conducted. 

\section{Conclusion and Outlook}\label{sec:conclusion}

In this paper, we presented \framework{}, an easy-to-use and versatile open source library for knowledge probing that can be seamlessly used with the \textsc{BEAR} probe.
The framework covers two important use cases: Monitoring knowledge during continual pre-training (and domain adaptation) and analyzing existing pre-trained language models.

We are actively working on extending the leaderboard of existing pre-trained language models and strongly encourage the community to participate. We aim to develop the library further to support other use cases and welcome any input, whether in the form of raised issues or contributions to the code base.

We are working to extend the BEAR probe to additional knowledge bases in order to expand on the domains of knowledge that can be evaluated with \framework{}.

\section*{Acknowledgements}

Max Ploner, Jacek Wiland, Sebastian Pohl, and Alan Akbik are supported by the Deutsche Forschungsgemeinschaft (DFG, German Research Foundation) under Germany’s Excellence Strategy – EXC 2002/1 “Science of Intelligence” – project number 390523135. Alan Akbik is further supported by the Deutsche Forschungsgemeinschaft (DFG, German Research Foundation) under the Emmy Noether grant ``Eidetic Representations of Natural Language'' (project number 448414230).

\bibliography{references}

\begin{thebibliography}{21}
\providecommand{\natexlab}[1]{#1}

\bibitem[{Clement et~al.(2019)Clement, Bierbaum, O'Keeffe, and
  Alemi}]{clement2019arxiv}
Colin~B. Clement, Matthew Bierbaum, Kevin~P. O'Keeffe, and Alexander~A. Alemi.
  2019.
\newblock \href {https://arxiv.org/abs/1905.00075} {On the use of arxiv as a
  dataset}.
\newblock \emph{Preprint}, arXiv:1905.00075.

\bibitem[{Cossu et~al.(2022)Cossu, Tuytelaars, Carta, Passaro, Lomonaco, and
  Bacciu}]{cossuContinualPreTrainingMitigates2022a}
Andrea Cossu, Tinne Tuytelaars, Antonio Carta, Lucia Passaro, Vincenzo
  Lomonaco, and Davide Bacciu. 2022.
\newblock \href {http://arxiv.org/abs/2205.09357} {Continual {Pre}-{Training}
  {Mitigates} {Forgetting} in {Language} and {Vision}}.
\newblock \emph{arXiv preprint}.
\newblock ArXiv:2205.09357 [cs].

\bibitem[{Devlin et~al.(2019)Devlin, Chang, Lee, and
  Toutanova}]{devlinBERTPretrainingDeep2019b}
Jacob Devlin, Ming-Wei Chang, Kenton Lee, and Kristina Toutanova. 2019.
\newblock \href {https://doi.org/10.18653/v1/N19-1423} {{BERT}: {Pre}-training
  of {Deep} {Bidirectional} {Transformers} for {Language} {Understanding}}.
\newblock In \emph{Proceedings of the 2019 {Conference} of the {North}
  {American} {Chapter} of the {Association} for {Computational} {Linguistics}:
  {Human} {Language} {Technologies}, {Volume} 1 ({Long} and {Short} {Papers})},
  pages 4171--4186, Minneapolis, Minnesota. Association for Computational
  Linguistics.

\bibitem[{Elsahar et~al.(2018)Elsahar, Vougiouklis, Remaci, Gravier, Hare,
  Laforest, and Simperl}]{elsaharTRExLargeScale2018}
Hady Elsahar, Pavlos Vougiouklis, Arslen Remaci, Christophe Gravier, Jonathon
  Hare, Frederique Laforest, and Elena Simperl. 2018.
\newblock \href {https://aclanthology.org/L18-1544} {T-{REx}: {A} {Large}
  {Scale} {Alignment} of {Natural} {Language} with {Knowledge} {Base}
  {Triples}}.
\newblock In \emph{Proceedings of the {Eleventh} {International} {Conference}
  on {Language} {Resources} and {Evaluation} ({LREC} 2018)}, Miyazaki, Japan.
  European Language Resources Association (ELRA).

\bibitem[{Gao et~al.(2023)Gao, Tow, Abbasi, Biderman, Black, DiPofi, Foster,
  Golding, Hsu, Le~Noac'h, Li, McDonell, Muennighoff, Ociepa, Phang, Reynolds,
  Schoelkopf, Skowron, Sutawika, Tang, Thite, Wang, Wang, and
  Zou}]{eval-harness}
Leo Gao, Jonathan Tow, Baber Abbasi, Stella Biderman, Sid Black, Anthony
  DiPofi, Charles Foster, Laurence Golding, Jeffrey Hsu, Alain Le~Noac'h,
  Haonan Li, Kyle McDonell, Niklas Muennighoff, Chris Ociepa, Jason Phang,
  Laria Reynolds, Hailey Schoelkopf, Aviya Skowron, Lintang Sutawika, Eric
  Tang, Anish Thite, Ben Wang, Kevin Wang, and Andy Zou. 2023.
\newblock \href {https://doi.org/10.5281/zenodo.10256836} {A framework for
  few-shot language model evaluation}.

\bibitem[{Geiger(2019)}]{geigerArXiVArchiveTidy2019}
R.~Stuart Geiger. 2019.
\newblock \href {http://doi.org/10.5281/zenodo.1463242} {{ArXiV} {Archive}: {A}
  {Tidy} and {Complete} {Archive} of {Metadata} for {Papers} on arxiv.org.}

\bibitem[{Haller et~al.(2024)Haller, Aynetdinov, and Akbik}]{opiniongpt2024}
Patrick Haller, Ansar Aynetdinov, and Alan Akbik. 2024.
\newblock \href {https://openreview.net/forum?id=Tv9lZ80yi6} {Opinion{GPT}:
  Modelling explicit biases in instruction-tuned {LLM}s}.
\newblock In \emph{2024 Annual Conference of the North American Chapter of the
  Association for Computational Linguistics -- System Demonstration Track}.

\bibitem[{Hendrycks et~al.(2021)Hendrycks, Burns, Basart, Zou, Mazeika, Song,
  and Steinhardt}]{hendryckstest2021}
Dan Hendrycks, Collin Burns, Steven Basart, Andy Zou, Mantas Mazeika, Dawn
  Song, and Jacob Steinhardt. 2021.
\newblock Measuring massive multitask language understanding.
\newblock \emph{Proceedings of the International Conference on Learning
  Representations (ICLR)}.

\bibitem[{Jiang et~al.(2023)Jiang, Sablayrolles, Mensch, Bamford, Chaplot,
  de~las Casas, Bressand, Lengyel, Lample, Saulnier, Lavaud, Lachaux, Stock,
  Scao, Lavril, Wang, Lacroix, and Sayed}]{jiang2023mistral7b}
Albert~Q. Jiang, Alexandre Sablayrolles, Arthur Mensch, Chris Bamford,
  Devendra~Singh Chaplot, Diego de~las Casas, Florian Bressand, Gianna Lengyel,
  Guillaume Lample, Lucile Saulnier, Lélio~Renard Lavaud, Marie-Anne Lachaux,
  Pierre Stock, Teven~Le Scao, Thibaut Lavril, Thomas Wang, Timothée Lacroix,
  and William~El Sayed. 2023.
\newblock \href {https://arxiv.org/abs/2310.06825} {Mistral 7b}.
\newblock \emph{Preprint}, arXiv:2310.06825.

\bibitem[{Kalo and Fichtel(2022)}]{kaloKAMELKnowledgeAnalysis2022}
Jan-Christoph Kalo and Leandra Fichtel. 2022.
\newblock \href
  {https://www.akbc.ws/2022/assets/pdfs/15_kamel_knowledge_analysis_with_.pdf}
  {{KAMEL} : {Knowledge} {Analysis} with {Multitoken} {Entities} in {Language}
  {Models}}.
\newblock In \emph{Automated {Knowledge} {Base} {Construction}}.

\bibitem[{Labs(2021)}]{bBritishLibraryBooks2021}
British~Library Labs. 2021.
\newblock Digitised books. c. 1510 - c. 1900. jsonl (ocr derived text +
  metadata).
\newblock https://doi.org/10.23636/r7w6-zy15.

\bibitem[{Liang et~al.(2023)Liang, Bommasani, Lee, Tsipras, Soylu, Yasunaga,
  Zhang, Narayanan, Wu, Kumar, Newman, Yuan, Yan, Zhang, Cosgrove, Manning,
  Ré, Acosta-Navas, Hudson, Zelikman, Durmus, Ladhak, Rong, Ren, Yao, Wang,
  Santhanam, Orr, Zheng, Yuksekgonul, Suzgun, Kim, Guha, Chatterji, Khattab,
  Henderson, Huang, Chi, Xie, Santurkar, Ganguli, Hashimoto, Icard, Zhang,
  Chaudhary, Wang, Li, Mai, Zhang, and
  Koreeda}]{liang2023holisticevaluationlanguagemodels}
Percy Liang, Rishi Bommasani, Tony Lee, Dimitris Tsipras, Dilara Soylu,
  Michihiro Yasunaga, Yian Zhang, Deepak Narayanan, Yuhuai Wu, Ananya Kumar,
  Benjamin Newman, Binhang Yuan, Bobby Yan, Ce~Zhang, Christian Cosgrove,
  Christopher~D. Manning, Christopher Ré, Diana Acosta-Navas, Drew~A. Hudson,
  Eric Zelikman, Esin Durmus, Faisal Ladhak, Frieda Rong, Hongyu Ren, Huaxiu
  Yao, Jue Wang, Keshav Santhanam, Laurel Orr, Lucia Zheng, Mert Yuksekgonul,
  Mirac Suzgun, Nathan Kim, Neel Guha, Niladri Chatterji, Omar Khattab, Peter
  Henderson, Qian Huang, Ryan Chi, Sang~Michael Xie, Shibani Santurkar, Surya
  Ganguli, Tatsunori Hashimoto, Thomas Icard, Tianyi Zhang, Vishrav Chaudhary,
  William Wang, Xuechen Li, Yifan Mai, Yuhui Zhang, and Yuta Koreeda. 2023.
\newblock \href {https://arxiv.org/abs/2211.09110} {Holistic evaluation of
  language models}.
\newblock \emph{Preprint}, arXiv:2211.09110.

\bibitem[{Liu et~al.(2019)Liu, Ott, Goyal, Du, Joshi, Chen, Levy, Lewis,
  Zettlemoyer, and Stoyanov}]{liuRoBERTaRobustlyOptimized2019}
Yinhan Liu, Myle Ott, Naman Goyal, Jingfei Du, Mandar Joshi, Danqi Chen, Omer
  Levy, Mike Lewis, Luke Zettlemoyer, and Veselin Stoyanov. 2019.
\newblock \href {https://doi.org/10.48550/arXiv.1907.11692} {{RoBERTa}: {A}
  {Robustly} {Optimized} {BERT} {Pretraining} {Approach}}.
\newblock \emph{arXiv preprint}.
\newblock ArXiv:1907.11692 [cs].

\bibitem[{Luo et~al.(2023)Luo, Vu, Phung, and Haffari}]{luo2023systematic}
Linhao Luo, Thuy-Trang Vu, Dinh Phung, and Gholamreza Haffari. 2023.
\newblock Systematic assessment of factual knowledge in large language models.
\newblock \emph{Findings of EMNLP}.

\bibitem[{McCloskey and Cohen(1989)}]{mccloskey1989catastrophic}
Michael McCloskey and Neal~J Cohen. 1989.
\newblock Catastrophic interference in connectionist networks: The sequential
  learning problem.
\newblock In \emph{Psychology of learning and motivation}, volume~24, pages
  109--165. Elsevier.

\bibitem[{Merity et~al.(2016)Merity, Xiong, Bradbury, and
  Socher}]{merity2016pointer}
Stephen Merity, Caiming Xiong, James Bradbury, and Richard Socher. 2016.
\newblock \href {https://arxiv.org/abs/1609.07843} {Pointer sentinel mixture
  models}.
\newblock \emph{Preprint}, arXiv:1609.07843.

\bibitem[{Petroni et~al.(2020)Petroni, Lewis, Piktus, Rocktäschel, Wu, Miller,
  and Riedel}]{petroniHowContextAffects2020}
Fabio Petroni, Patrick Lewis, Aleksandra Piktus, Tim Rocktäschel, Yuxiang Wu,
  Alexander~H. Miller, and Sebastian Riedel. 2020.
\newblock \href {http://arxiv.org/abs/2005.04611} {How {Context} {Affects}
  {Language} {Models}' {Factual} {Predictions}}.
\newblock \emph{arXiv preprint}.
\newblock ArXiv:2005.04611 [cs].

\bibitem[{Petroni et~al.(2019)Petroni, Rocktäschel, Riedel, Lewis, Bakhtin,
  Wu, and Miller}]{petroniLanguageModelsKnowledge2019}
Fabio Petroni, Tim Rocktäschel, Sebastian Riedel, Patrick Lewis, Anton
  Bakhtin, Yuxiang Wu, and Alexander Miller. 2019.
\newblock \href {https://doi.org/10.18653/v1/D19-1250} {Language {Models} as
  {Knowledge} {Bases}?}
\newblock In \emph{Proceedings of the 2019 {Conference} on {Empirical}
  {Methods} in {Natural} {Language} {Processing} and the 9th {International}
  {Joint} {Conference} on {Natural} {Language} {Processing}
  ({EMNLP}-{IJCNLP})}, pages 2463--2473, Hong Kong, China. Association for
  Computational Linguistics.

\bibitem[{Radford et~al.(2019)Radford, Wu, Child, Luan, Amodei, Sutskever
  et~al.}]{radfordLanguageModelsAre2019}
Alec Radford, Jeffrey Wu, Rewon Child, David Luan, Dario Amodei, Ilya
  Sutskever, et~al. 2019.
\newblock Language models are unsupervised multitask learners.
\newblock \emph{OpenAI blog}, 1(8):9.

\bibitem[{Wiland et~al.(2024)Wiland, Ploner, and Akbik}]{wilandBEAR2024}
Jacek Wiland, Max Ploner, and Alan Akbik. 2024.
\newblock \href {https://aclanthology.org/2024.findings-naacl.155} {{BEAR}: A
  unified framework for evaluating relational knowledge in causal and masked
  language models}.
\newblock In \emph{Findings of the Association for Computational Linguistics:
  NAACL 2024}, pages 2393--2411, Mexico City, Mexico. Association for
  Computational Linguistics.

\bibitem[{Wolf et~al.(2020)Wolf, Debut, Sanh, Chaumond, Delangue, Moi, Cistac,
  Rault, Louf, Funtowicz, Davison, Shleifer, von Platen, Ma, Jernite, Plu, Xu,
  Le~Scao, Gugger, Drame, Lhoest, and
  Rush}]{wolfTransformersStateoftheArtNatural2020}
Thomas Wolf, Lysandre Debut, Victor Sanh, Julien Chaumond, Clement Delangue,
  Anthony Moi, Pierric Cistac, Tim Rault, Remi Louf, Morgan Funtowicz, Joe
  Davison, Sam Shleifer, Patrick von Platen, Clara Ma, Yacine Jernite, Julien
  Plu, Canwen Xu, Teven Le~Scao, Sylvain Gugger, Mariama Drame, Quentin Lhoest,
  and Alexander Rush. 2020.
\newblock \href {https://doi.org/10.18653/v1/2020.emnlp-demos.6} {Transformers:
  {State}-of-the-{Art} {Natural} {Language} {Processing}}.
\newblock In \emph{Proceedings of the 2020 {Conference} on {Empirical}
  {Methods} in {Natural} {Language} {Processing}: {System} {Demonstrations}},
  pages 38--45, Online. Association for Computational Linguistics.

\end{thebibliography}

\clearpage
\appendix

\section{Additional Information on the Experiments}

Due to the number of experiments and the limited space, we provide additional information on the experiments we presented in this part of the appendix.

\subsection{Continual Pre-training Experiment}

\paragraph{Dataset}
In our experiments with continual learning of the \texttt{bert-base-cased} in the Section \ref{sec:cp-experiment}, we use a subset of the arXiv dataset \citep{geigerArXiVArchiveTidy2019}. We use the same data splits as \citet{cossuContinualPreTrainingMitigates2022a}, i.e., same document classes and observations. Specifically, the following classes of scientific abstracts were used: \texttt{`hep-ph'}, \texttt{`astro-ph'}, \texttt{`hep-th'}, \texttt{`quant-ph'}, \texttt{`cond-mat.mes-hall'}, \texttt{`gr-qc'}, \texttt{`cond-mat.mtrl-sci'}, \texttt{`cond-mat.str-el'}, \texttt{`condmat.stat-mech'} and \texttt{`astro-ph.SR'}. 
Selecting these specific abstracts enables us to evaluate their findings on mitigating forgetting during self-supervised learning. These scientific domains primarily span physics and materials science. Each of these ten classes has a training set of approximately 10,000 abstracts and a validation set of about 1,000 abstracts.

\paragraph{Training hyperparameters}

The hyperparameters used during reported in Table~\ref{tab:hyperparameters}.

\begin{table}[h!]
\centering
\scriptsize
\begin{tabularx}{\linewidth}{X|c}
\toprule
\textbf{Hyperparameter} & \textbf{Value} \\
\midrule
Per Device Train Batch Size & 8 \\
Per Device Eval Batch Size & 8 \\
Gradient Accumulation Steps & 1 \\
Learning Rate & 0.00005 \\
Weight Decay & 0 \\
Number of Training Epochs & 30 \\
Learning Rate Scheduler Type & Linear \\
Warmup Ratio & 0.0 \\
Metric for Best Model & Evaluation Loss \\
Early Stopping Patience & 5 \\ 
Early Stopping Threshold & 0 \\
\bottomrule
\end{tabularx}
\caption{Hyperparameters used during continual pre-training.}
\label{tab:hyperparameters}
\end{table}

\paragraph{Additional Results \& Discussion}\label{sec:cl-experiment-additional}
The relative performance of \texttt{bert-base-cased} measured on T-REx task after the \textit{ith} experience of continual pre-training as measured by the \textsc{LM Pub Quiz} and [MASK]-predict are shown in Table~\ref{tab:continous_pretrainig_experiment2}. The scores are normalized with respect to their base performance before continual pre-training. \textit{0th} experience corresponds to the original model taken from Hugging Face.

The [MASK]-predict technique exhibits a significant degradation in performance from the outset of continual pre-training, with over a 75\% decrease observed after the first experience. Overall, this method suggests that nearly 95\% of the knowledge was lost training on the arXiv dataset. On the other hand, results obtained with \textsc{LM Pub Quiz} show a relatively smaller decrease of approximately 60\%.

A certain degree of this difference in degradation can be explained by the difference in random baseline. When using [MASK]-predict, degrading to the level of the random baseline would amount to a drop of almost 100\% while when using \textsc{LM Pub Quiz} this would lead to a drop of only roughly 90\% due to a higher accuracy of the random baseline (given the smaller answer space).

\begin{table}[t]
  \centering
  \scriptsize
    \begin{tabularx}{\linewidth}{>{\centering}Xcc}
    \toprule
\makecell{\textbf{Evaluation /} \\ \textbf{Experience}} &  \makecell{\textbf{T-REx} \\ \textbf{[MASK]-predict (\%)}} & \makecell{\textbf{T-REx} \\ \textbf{LM Pub Quiz (\%)}}  \\
    \midrule
          0 &                100.00 &             100.00  \\
          1 &                 24.12 &              72.37 \\
          2 &                  9.98 &              50.09  \\
          3 &                  3.79 &              39.04  \\
          4 &                  4.41 &              38.58  \\
    \bottomrule
  \end{tabularx}
  \vspace*{-0.2cm}
  \caption{{The relative performance of \texttt{bert-base-cased} measured on T-REx task during continual pre-training.
  Before the continual pre-training the model achieves 31.3\% accuracy using [MASK]-predict and 40.5\% using \textsc{LM Pub Quiz} as well as 18.4\% accuracy on BEAR.
  }}
  \label{tab:continous_pretrainig_experiment2}
  \vspace*{-0.2cm}
\end{table}

\subsection{Training on Different Domain Corpora}\label{sec:continued-domain-specific-knowledge}
\begin{table}[h!]
\centering
\footnotesize
\begin{tabularx}{\linewidth}{X|c}
\toprule
\textbf{Domain} & \textbf{Num. of Train Tokens} \\
\midrule
arXiv & $2,49e^8$ \\
blbooks & $1,72e^8$ \\
wikitext & $2,82e^8$ \\
\bottomrule
\end{tabularx}
\caption{The number of tokens seen by each individual adapted model. The \texttt{wikitext-103-v1} dataset contained this number of tokens in total after some minor cleaning.}
\label{tab:dataset-sizes}
\end{table}

Each model was trained on a similar number of tokens (see Table~\ref{tab:dataset-sizes}). We trained four models per dataset over random permutations of the data. The arXiv dataset used was split into four equal chunks of the given size. The blbooks dataset was split into more chunks, but only four chunks of the given size were used for the training of four models. All models were trained with the  hyperparameters reported in Table~\ref{tab:domain-hyperparameters}.

\begin{table}[h!]
\centering
\scriptsize
\begin{tabularx}{\linewidth}{X|c}
\toprule
\textbf{Hyperparameter} & \textbf{Value} \\
\midrule
Per Device Train Batch Size & 32 \\
Gradient Accumulation Steps & 1 \\
Learning Rate & 1e-05 \\
Weight Decay & 0 \\
Number of Training Epochs & 1 \\
Learning Rate Scheduler Type & Cosine \\
Warmup Ratio & 0.0 \\
\bottomrule
\end{tabularx}
\caption{Hyperparameters Used for Model Training on Different Domains}
\label{tab:domain-hyperparameters}
\end{table}

\subsection{Pre-trained Model Bias}\label{sec:continued-model-bias}

\begin{figure}[ht]
    \centering
    \includegraphics[width=0.95\columnwidth]{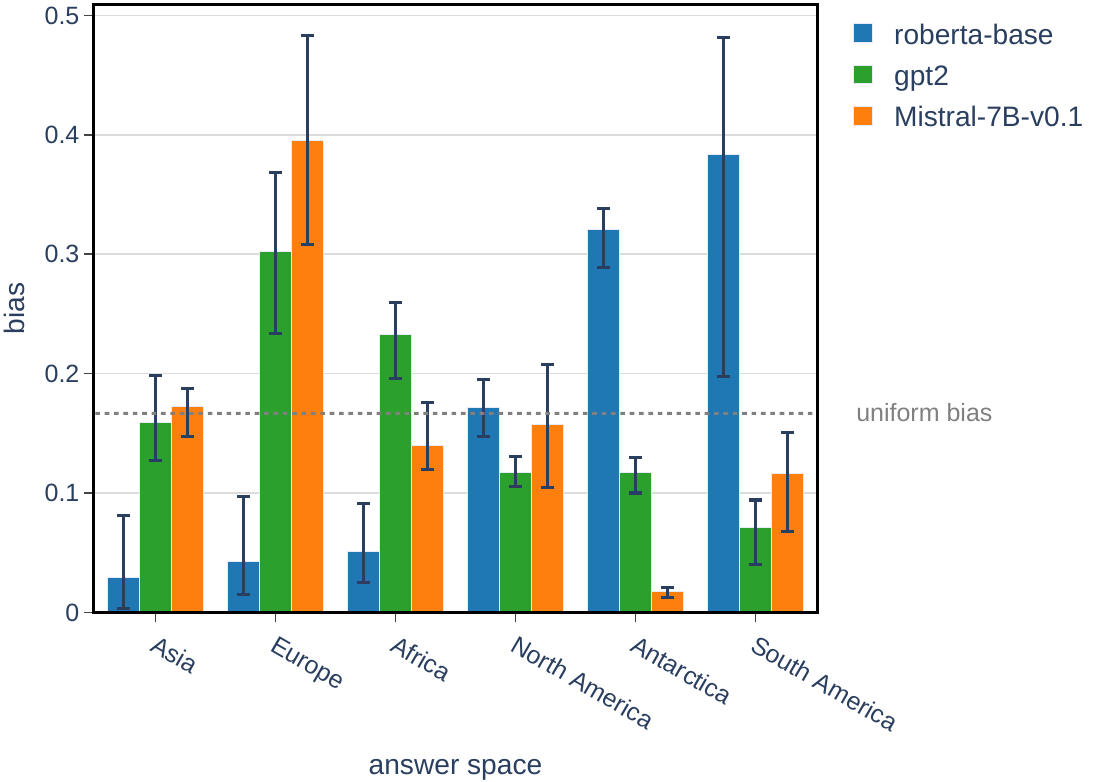}
    \caption{Generic subject biases for relation \texttt{P30} of the \textsc{BEAR} probe for various models.}\label{fig:bias_generic_p30}
\end{figure}

Extending the results of section \ref{sec:model-bias} we also estimated model biases for relation \texttt{P30} using only six manually chosen generic subjects for the relation, including for example `it` and `the region`.  Model biases are again estimated by applying Softmax to the BEAR pseudo-log-likelihood scores and by averaging the resulting distributions over all generic subjects.

Figure \ref{fig:bias_generic_p30} shows the results for this way of calculating biases. We can observe the same trend mentioned before: \texttt{roberta-base} is biased towards South America and Antarctica, whereas \texttt{GPT2} and \texttt{Mistral-7B-v0.1} are biased towards Europe. But this time \texttt{Mistral-7B-v0.1} appears to be even more biased than \texttt{GPT2}. When biases are computed in this second way, they indicate, which answers a model chooses without subject information. While \texttt{Mistral-7B-v0.1} shows a high bias here, it still predicts many correct answers resulting in a lower bias according to the first method. It appears as though its subject-specific knowledge overcomes this bias, while the smaller less performative \texttt{GPT2} is less able to overcome this kind of bias.

\section{Additional Information on the Use of other Datasets}\label{sec:dataset-building}

The package was primarily developed to enable the use of the BEAR dataset. Still, the approach is quite general and can be used to cover other domains than the rather general knowledge represented in this subset of wikidata or answer different research questions altogether.

Each dataset consists of a set of relations (JSONL files)  and metadata in the \texttt{metadata\_relations.json} JSON file. For each relation, the metadata file contains one or more templates and (optionally) the definition of an answer space (see Listing~\ref{lst:dataset-def}). If no answer space is given, it is constructed from all objects in the relation.

\begin{figure}[ht]
    \centering
    \begin{lstlisting}[caption=Definiton of the templates and answer spaces of relations in \texttt{metadata\_relations.json} of a \textsc{LM Pub Quiz} dataset.,label={lst:dataset-def},language=json]{Dataset Definition}
{
    "<relation id>": {
        "templates": [
            "[Y] is the answer to some fact with subject [X].",
            ...
        ],
        "answer_space_labels": [
            "<some object label>",
            ...
        ],
        "answer_space_ids": [
            "<object ids>",
            ...
        ]
    },
    ...
}
    \end{lstlisting}
\end{figure}

The templates are used to construct alternative textual statements: ``[X]'' is replaced by the subject of the instance and ``[Y]'' is replaced by each of the options in the answer space to construct one statement per answer option.

Each relation contains one instance per line (the file should be named \texttt{<relation id>.jsonl}; see Listing~\ref{lst:instance-def}).
Each  (represented as by a JSON object) should have a subject and object (i.e. the correct answer) ID as well as labels for the subject (and object if the answer space is not defined in the metadata).\footnote{The IDs of the subjects and objects should be unique (though they can be shared across the relations) and may refer to the IDs of the underlying knowledge base. Additional fields (such as aliases) are not used at the moment.}
The instances require either an object ID or the index of the correct answer in the answer space defined in the metadata file.

\begin{figure}[ht]
    \centering
    \begin{lstlisting}[caption=Definiton of instances in a relation of a \textsc{LM Pub Quiz} dataset (single line).,label={lst:instance-def},language=json]{Instance Definition}
{"sub_id": "<subject id>", "sub_label": "<subject label>", "obj_id": "<ID of the correct answer>", "obj_label": "<correct object label>", "answer_idx": <index of the correct object in the answer space>}
    \end{lstlisting}
\end{figure}

\end{document}